\documentclass[10pt, conference, compsocconf]{IEEEtran}

\usepackage{url}
\usepackage{cite}
\usepackage{amsmath,amssymb,amsfonts}
\usepackage{algorithmic}
\usepackage{graphicx}
\usepackage{textcomp}
\usepackage{xcolor}
\usepackage{subcaption}
\usepackage{listings}
\usepackage{xcolor}
\usepackage{booktabs}

\definecolor{background}{RGB}{250,250,250}
\definecolor{numb}{RGB}{220,0,0}
\definecolor{punct}{RGB}{0,0,0}
\definecolor{delim}{RGB}{0,0,255}

\newcommand{\ie}{\textit{i}.\textit{e}., }
\newcommand{\eg}{\textit{e}.\textit{g}., }

\newcommand{\cg}{\mathrm{CG}}
\newcommand{\tg}{\mathrm{TG}}
\newcommand{\knn}{\mathrm{kNN}}
\newcommand{\dist}{\mathrm{dist}}
\newcommand{\gnn}{\mathrm{GNN}}
\newcommand{\cnn}{\mathrm{CNN}}

\newcommand{\roi}{\mathrm{RoI}}
\newcommand{\hact}{\mathrm{HACT}}
\newcommand{\concat}{\mathrm{Concat}}
\newcommand{\cgc}{\mathrm{CGCNet}}
\newcommand{\gin}{\mathrm{GIN}}

\usepackage{hyperref}
\hypersetup{
    colorlinks   = true,
    citecolor    = green
}

\begin{document}

\title{HACT-Net: A Hierarchical Cell-to-Tissue Graph Neural Network \\ for Histopathological Image Classification}


% \author{\IEEEauthorblockN{Guillaume Jaume\footnote{Corresponding author}}
% \IEEEauthorblockA{Swiss Federal Institute of Technology\\
% Signal Processing Laboratory 5 \\ 
% Lausanne, Switzerland \\
% guillaume.jaume@epfl.ch}
% \and 
% \IEEEauthorblockN{Haz{\i}m Kemal Ekenel}
% \IEEEauthorblockA{Istanbul Technical University\\
% Department of Computer Engineering \\
% Istanbul, Turkey \\
% ekenel@itu.edu.tr}
% \and
% \IEEEauthorblockN{Jean-Philippe Thiran}
% \IEEEauthorblockA{Swiss Federal Institute of Technology\\
% Signal Processing Laboratory 5 \\ 
% Lausanne, Switzerland \\
% jean-philippe.thiran@epfl.ch}
% }

\author{
\IEEEauthorblockN{Pushpak Pati*$^{1,2}$}
\and
\IEEEauthorblockN{Guillaume Jaume*$^{2,3}$}
\and 
\IEEEauthorblockN{Lauren Alisha Fernandes$^{1}$}
\and
\IEEEauthorblockN{Antonio Foncubierta$^{3}$}
\and
\IEEEauthorblockN{Florinda Feroce$^{4}$}
\and
\IEEEauthorblockN{Anna Maria Anniciello$^{4}$}
\and
\IEEEauthorblockN{Giosue Scognamiglio$^{4}$}
\and
\IEEEauthorblockN{Nadia Brancati$^{5}$}
\and
\IEEEauthorblockN{Daniel Riccio$^{5}$}
\and
\IEEEauthorblockN{Maurizio Do Bonito$^{4}$}
\and
\IEEEauthorblockN{Giuseppe De Pietro$^{5}$}
\and
\IEEEauthorblockN{Gerardo Botti$^{4}$}
\IEEEauthorblockA{\\$^1$ETH Zurich\\
Zurich, Switzerland} 
\and
\IEEEauthorblockN{Orcun Goksel$^{1}$}
\IEEEauthorblockA{\\$^2$EPFL\\
Lausanne, Switzerland}
\and
\IEEEauthorblockN{Jean-Philippe Thiran$^{2}$}
\IEEEauthorblockA{\\$^3$IBM Research\\
Zurich, Switzerland}
\and
\IEEEauthorblockN{Maria Frucci$^{5}$}
\IEEEauthorblockA{\\$^4$IRCCS-Fondazione Pascale\\
Naples, Italy}
\and
\IEEEauthorblockN{Maria Gabrani$^{3}$}
\IEEEauthorblockA{\\$^5$ICAR-CNR\\
Naples, Italy}
}


\maketitle

\begin{abstract}
Cancer diagnosis, prognosis, and therapeutic response prediction are heavily influenced by the relationship between the histopathological structures and the function of the tissue.
Recent approaches acknowledging the structure-function relationship, have linked the structural and spatial patterns of cell organization in tissue via cell-graphs to tumor grades. Though cell organization is imperative, it is insufficient to entirely represent the histopathological structure.
% PROPOSED METHOD
We propose a novel hierarchical cell-to-tissue-graph (HACT) representation
%, that mimics a pathologist's diagnostic procedure, 
to improve the structural depiction of the tissue. 
It consists of a low-level cell-graph, capturing cell morphology and interactions, a high-level tissue-graph, capturing morphology and spatial distribution of tissue parts, and cells-to-tissue hierarchies, encoding the relative spatial distribution of the cells with respect to the tissue distribution.
Further, a hierarchical graph neural network (HACT-Net) is proposed to efficiently map the HACT representations to histopathological breast cancer subtypes.
% DATASET AND EVALUATION
We assess the methodology on a large set of annotated tissue regions of interest from H\&E stained breast carcinoma whole-slides. 
%The dataset consists of challenging typical and atypical subtypes, and incorporate realistic diagnostic issues.
Upon evaluation, the proposed method outperformed recent convolutional neural network and graph neural network approaches for breast cancer multi-class subtyping.
% BRIEF CONCLUSION
The proposed entity-based topological analysis is more inline with the pathological diagnostic procedure of the tissue. It provides more command over the tissue modelling, therefore encourages the further inclusion of pathological priors into task-specific tissue representation. 

%It encourages further inclusion of pathological priors into tissue modeling. Additionally, our methodology paves way for the interpretability techniques in digital pathology to explore the hierarchical nature of tissue. \gja{remove the interpretability part}
%Our results encourage further inclusion of clinical diagnostic aspects into the hierarchical assessment of the tissue.

% Cancer diagnosis, prognosis and therapeutic response prediction are heavily influenced by the relationship between the histopathological structures and the function of the tissue. 
\end{abstract}


% \fi
\IEEEpeerreviewmaketitle

%------------------------------------------------------------------
% INTRODUCTION (RELATED WORK INCLUDED)
%------------------------------------------------------------------
\section{Introduction}
\label{introduction}

Breast cancer is the second most common type of cancer with high mortality rate in women \cite{siegel16}.
A majority of breast lesions are diagnosed according to a diagnostic spectrum of cancer classes that ranges from benign to invasive. The classes confer different folds of risk to become invasive. 
Lesions with atypia or ductal carcinoma in-situ are associated with higher risks of transitioning to invasive carcinoma compared to benign lesions\cite{myers19,Elmore2015}.
Thus, accurate discrimination of these classes is pivotal to determine the optimal treatment plan. 
However, distinguishing the classes is not always easy, \eg in~\cite{Elmore2015}  pathologists' concordance rates were as low as 48\% for atypia.
In a clinical setting, pathologists begin the classification of a tissue biopsy by discerning the morphology and the spatial distribution of tissue parts, such as epithelium, stroma, necrosis etc. 
Then, they localize their analysis to specific regions of interest ($\roi$) on the tissue and evaluate nuclear phenotype, morphology, topology and tissue distribution among several other criteria for the classification. However, such inspections are tedious, time-consuming and prone to observer variability, thus increasing the demand for automated systems in cancer diagnosis. 
%Due to the increased incidence of breast cancer and aforementioned challenges, there is an increasing demand for automated systems in cancer diagnosis. 
%The automatic systems can be used to sieve out obvious normal/benign slides to facilitate pathologists in analyzing more important abnormalities.
Digital pathology has recently motivated innovative research opportunities in machine learning and computer vision to automate cancer diagnosis \cite{litjens17}.
The most common technique for classifying $\roi$s consists of extracting fixed-size patches from an $\roi$ and classifying them using Convolutional Neural Networks ($\cnn$); then, patch-based predictions are aggregated to label the $\roi$ \cite{mercan19, aresta19}.
Such approaches are limited to finding the apt patch size and resolution to include context information.
It can be achieved by reducing the resolution at the cost of missing cell-level information, or by increasing the resolution at the cost of limiting patch size due to computational challenges. 
Additionally, patch-based approaches unfairly assume the same label for an $\roi$ and its corresponding patches.
Further, the pixel-based analysis by the CNNs do not comprehend the essence of biological entities and their biological context. This inhibits the integration of CNNs and prior pathological knowledge that would require selective entity-based application of CNNs.
%Finally, CNNs applied without biological context make the learned representations harder to interpret with respect to the tissue, thus limiting their integration with prior pathological knowledge that would require selective application of CNNs depending on tissue attributes.

\iffalse
However, this approach incurs several drawbacks.
1) Patches contain partial information about the $\roi$. Thus, assuming same label for all the patches corresponding to a $\roi$~results in noisy training. 
2) Balancing between patch size and operating resolution is cumbersome. A lower resolution allows for capturing global $\roi$~context, but limits the local cell-level information. Whereas, operating at a higher resolution provides desired cell-level information, but limits the patch size due to computational bottleneck, thus limiting the global context. 
3) Considering a high variance in $\roi$~dimensions, working with fixed size patches may include undesired regions or may exclude vital context information.
4) Convolutions operate at the pixel-level without meaningfully considering the biological context of the pixel. Thus, the patch-wise learned representations lack interpretable correspondence to the tissue attributes.
5) Further, CNNs cannot be selectively employed on tissue structures based on pathological understanding of the tissue, thus limiting its integration with prior pathological knowledge.
\fi

To address the above issues, histopathological structures of tissues have been represented by cell-graphs ($\cg$) \cite{gunduz04}, where cells and cellular interactions are presented as nodes and edges of $\cg$ respectively. Then, classical graph learning techniques or graph neural networks ($\gnn$s) learn from $\cg$s to map the structure-function relationship.
Recently various $\cg$ representations \cite{sharma16 , gadiya19 , zhou19, wang19tma} have been proposed by varying the graph building strategies or the node attributes.
However, a $\cg$ exploits only the cellular morphology and topology, and discards the tissue distribution information such as the stromal microenvironment, tumor microenvironment, lumen structure etc. that are vital for appropriate representation of histopathological structures. 
Additionally, a $\cg$ cannot represent the hierarchical nature of the tissue.
For instance, in \cite{zhou19}, a hierarchy is defined from the cells with learned pooling layers. However, the tissue hierarchy is inaccessible as the representation does not include high-level tissue features.
In \cite{chen19}, the cell-level and tissue-level information are simply concatenated. Thus, the functional representation of the tissue cannot leverage the hierarchy between the levels.

We address the above shortcomings by proposing a novel HierArchical-Cell-to-Tissue ($\hact$) representation of the $\roi$s. In $\hact$ representation, a low-level $\cg$~captures the cellular morphology and topology; a high-level tissue-graph ($\tg$) captures the attributes of the tissue parts and their spatial distribution; and the hierarchy between the $\cg$~and the $\tg$~captures the relative distribution of the cells with respect to the tissue distribution.
Further, we propose HACT-Net, a hierarchical $\gnn$ to learn from the $\hact$~representation and predict cancer types. Similar to the $\roi$ diagnostic procedure by the pathologist's, HACT-Net encodes contextual local and global structural attributes and interactions, thereby allowing for enriched structure-function relation analysis.

\iffalse
\section{Related work}
\label{related_work}
\subsection{Graph Neural Networks}
Graph Neural Networks (GNN) define a class of neural networks that are, by design, able to operate on graph-structured data.
They extend the concept of convolution that is well defined on regular structures, \eg on images, to irregular domains. Therefore, GNN can be regarded as a generalization of convolutional neural networks (ConvNets).
More specifically, in this paper we use GNNs that are based on message passing. In message passing neural networks (MPNN), each node is initially represented by a state, \ie features that are describing it. The node states are then iteratively updated following two steps. In an aggregation step, the neighboring node states are pooled into a single representation using a differentiable and permutation-invariant operator, \eg a sum. Then, in an update step, a new state for each node is computed by applying a differentiable operator, \eg a multi-layer perceptron (MLP), to the current node state and to the aggregated vector. After $N$ iterations, in order to build a fixed-size graph-level representation, all the node states are pooled with a differentiable permutation-invariant readout function. 
\fi

\begin{figure*}
    \includegraphics[width=\linewidth]{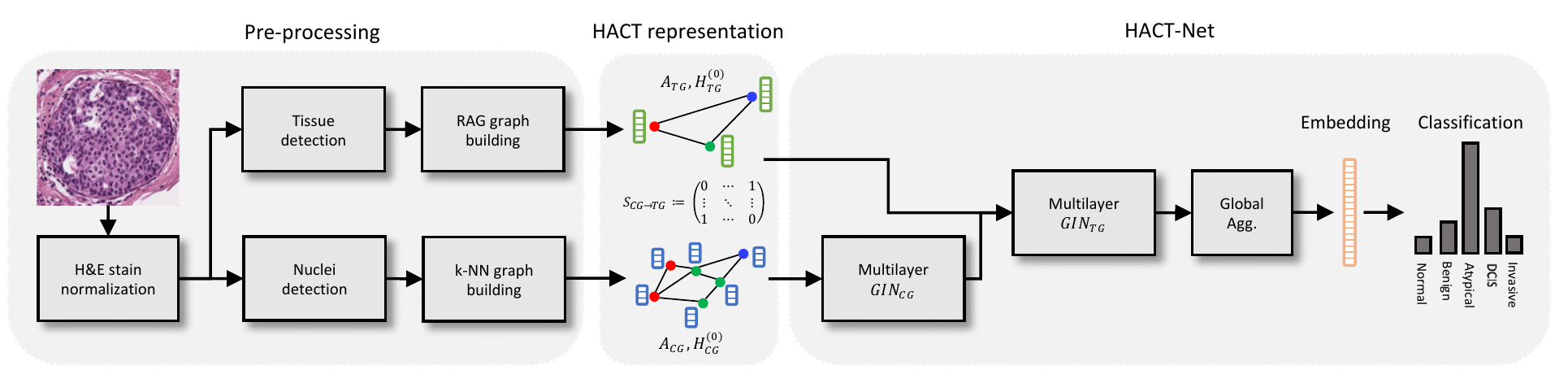}
    \caption{Block diagram of the proposed methodology including pre-processing module, HACT representation of a $\roi$ and HACT-Net classifying the $\roi$.}
    \label{fig:block_diagram}
\end{figure*}

\section{Methods}
\label{proposed_methodology}
We propose a $\hact$-representation that consists of a low-level $\cg$, a high level $\tg$ and cell-to-tissue hierarchies.
This representation is processed by $\hact$-Net, a hierarchical GNN that employs two GNNs~\cite{deferrard16, hamilton17, kipf17, velikovic18, xu19, gilmer17} to operate at cell and tissue-level.
The learned cell \emph{node} embeddings are combined with the corresponding tissue \emph{node} embedding via the cell-to-tissue hierarchies.
Figure~\ref{fig:block_diagram} summarizes the proposed methodology including the pre-processing for stain normalization~\cite{macenko09}, $\hact$-representation building and $\hact$-Net.

\subsection{Representation}
We define an undirected graph $G := (V, E)$ as a set of $|V|$ nodes and $|E|$ edges. An edge between the nodes $u$ and $v$ is denoted by $e_{uv}$ or $e_{vu}$.
The graph topology is described by a symmetric adjacency matrix $A \in \mathbb{R}^{|V| \times |V|}$, where an entry $A_{u,v} = 1$ if $e_{uv} \in E$.
Each node $v$ is presented by a feature vector $h(v) \in \mathbb{R}^d$.
Equivalently, the node features are presented in their matrix form as $H \in \mathbb{R}^{|V| \times d}$. 
We define the neighborhood of a node $v$ as $\mathcal{N}(v) := \{u \in V \; | \; v \in V, \; e_{uv} \in E \;\}$. 
%Note that in the following analysis, all the graphs are assumed to be undirected, \ie $e_{u \rightarrow v} \in E \implies e_{v \rightarrow u} \in E$. 
\\
\subsubsection{Cell-graph ($\cg$)}
In a $\cg$, each node represents a cell and edges encode cellular interactions. We detect nuclei using the Hover-Net model \cite{graham19}, pre-trained on the multi-organ nuclei segmentation dataset \cite{kumar17}.
%a state-of-the-art nuclei segmentation algorithm based on fully convolutional neural networks, 
For each detected nucleus at 40$\times$ resolution, we extract hand-crafted features representing shape, texture and spatial location following \cite{zhou19}.
Shape features include eccentricity, area, maximum and minimum length of axis, perimeter, solidity and orientation. Texture features include average foreground and background difference, standard deviation, skewness and mean entropy of nuclei intensity, and dissimilarity, homogeneity, energy and ASM from Gray-Level Co-occurrence Matrix. 
Nuclei are spatially encoded by their spatial centroids normalised by the image size. In total, each nucleus is represented by 18 features, noted as $f_\cg$. These features serve as the initial node embeddings in $\cg$.

To generate the $\cg$~topology, we assume that spatially close cells encode biological interactions and should be connected in $\cg$, and distant cells have weak cellular interactions, so they should remain disconnected in $\cg$.
To this end, we use the k-Nearest Neighbors ($\knn$) algorithm to build the initial topology, and prune the $\knn$~graph by removing edges lengthier than a threshold distance $d_{min}$. We use $L2$ norm in the image space to quantify the cellular distance. 
Formally, for each node $v$, an edge $e_{vu}$ is built if $u \in \{ w \; | \; \dist(v, w) \leq d_k \wedge \dist(v, w) < d_{\min}, \; \forall w \in V, \; v \in V, d_k= k\text{-th smallest distance in } \dist(v, w)\}$.
In our experiments, we set $k=5$ and $d_{min}=50$ pixels, i.e. 12.5 $\mu$m considering the scanner resolution of 0.25 $\mu$m/pixel.
Figure~\ref{fig:hierarchical_graph}\textcolor{red}{(a)} presents a sample $\cg$ elucidating the nodes and edges in the zoomed-in sub-image.
\\
\subsubsection{Tissue-graph ($\tg$)}
To capture the tissue distribution, we construct a $\tg$ by considering interactions among the parts of the tissue. In particular, we consider the SLIC algorithm~\cite{achanta11} emphasizing on space proximity to over-segment tissue parts into non-overlapping homogeneous super-pixels.
Subsequently, to create super-pixels capturing meaningful tissue information, we hierarchically merge adjacent similar super-pixels. The similarity is measured by texture attributes, \ie contrast, dissimilarity, homogeneity, energy, entropy and ASM from Gray-Level Co-occurrence Matrix, and channel-wise color attributes, \ie 8-bin color histogram, mean, standard deviation, median, energy and skewness.
Initial over-segmentation is performed at $10\times$ magnification to detect more homogeneous super-pixels and to achieve computational efficiency in super-pixel detection.
Finally, color and texture features are extracted for the merged super-pixels at $40\times$ magnification to capture informative local attributes. A supervised random-forest feature selection is employed and 24 dominant features are selected that classify the super-pixels into epithelium, stroma, necrosis and background tissue parts. Additionally, spatial centroids of super-pixels normalised by the image size are included to construct 26-dimensional representations for the super-pixels.

To generate the $\tg$~topology, we assume that adjacent tissue parts biologically interact and should be connected. To this end, we construct a region adjacency graph (RAG) \cite{potjer96} using the spatial centroids of the super-pixels. The super-pixel attributes define the initial node features, noted as $f_\tg$ and the RAG edges define the $\tg$ edges. Figure~\ref{fig:hierarchical_graph}\textcolor{red}{(b)} presents a sample $\tg$. The large node at the center represents the centroid of the surrounding stroma that is connected to the parts of epithelium and background. Thus, $\tg$ encodes information from the tumor and the stroma microenvironment.
\\
\subsubsection{HierArchical-Cell-to-Tissue ($\hact$) representation}
\begin{figure}[t]
  \centering
  \begin{minipage}{.25\linewidth}
    \centering
    \subcaptionbox{}
      {\includegraphics[width=\linewidth]{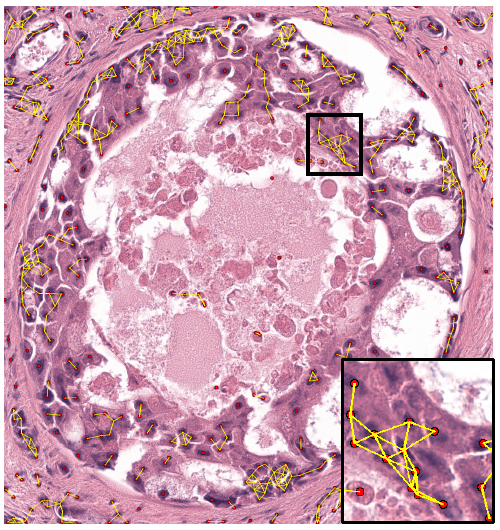}}

    \subcaptionbox{}
      {\includegraphics[width=\linewidth]{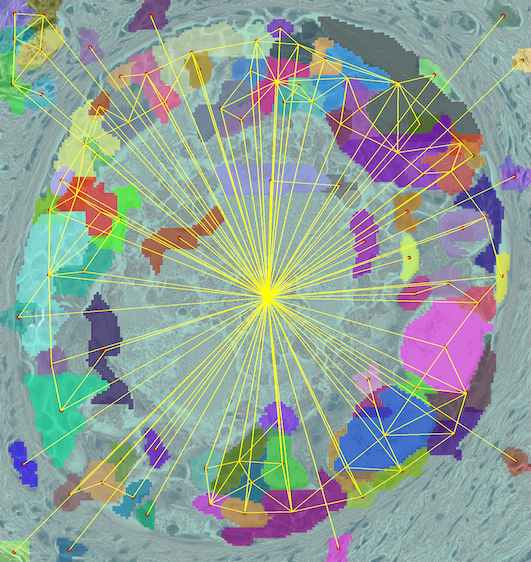}}

  \end{minipage}\quad
  \begin{minipage}{.55\linewidth}
    \centering
    \subcaptionbox{}
      {\includegraphics[width=\linewidth]{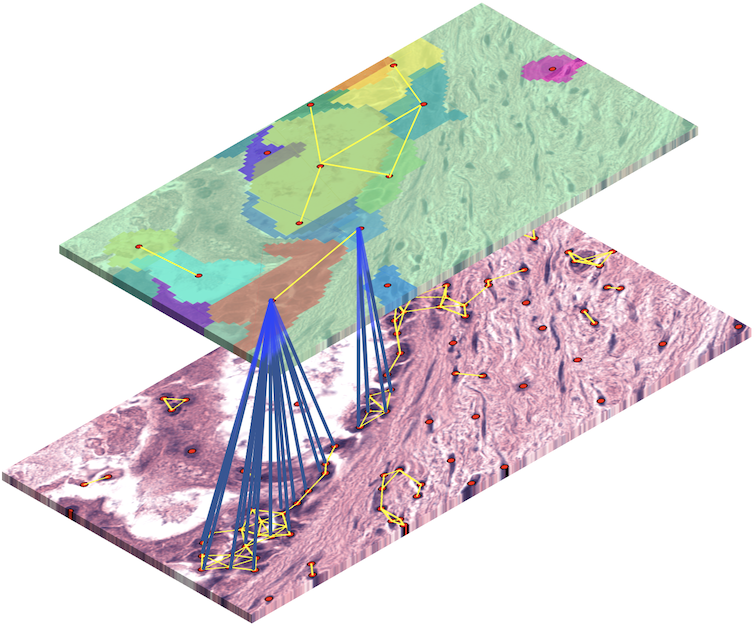}}
  \end{minipage}
 \caption{Visualizing (a) $\cg$, (b) $\tg$, and (c) $\hact$ representations. Nodes are presented in red and edges in yellow. Cell-to-tissue hierarchies are shown in blue in $\hact$. Note that all hierarchies in $\hact$ are not shown for visual clarity.}
 \label{fig:hierarchical_graph}
 \end{figure}

To jointly represent the low-level $\cg$ and high-level $\tg$, we introduce $\hact$ defined as $G_\hact := \{G_\cg, G_\tg, S_{\cg \rightarrow \tg}\}$. $G_\cg  = (V_\cg, E_\cg)$ and $G_\tg = (V_\tg, E_\tg)$ are $\cg$ and $\tg$ respectively.
We introduce an assignment matrix $S_{\cg \rightarrow \tg} \in \mathbb{R}^{|V_\cg| \times |V_\tg|}$ that describes a pooling operation to topologically map $\cg$ to $\tg$.
$S_{\cg \rightarrow \tg}$  is built using the spatial information of nuclei and super-pixels, \ie $S_{\cg \rightarrow \tg}(i, j) = 1$ if the nucleus represented by node $i$ in $\cg$ spatially belongs to the super-pixel represented by node $j$ in $\tg$.
Note that $|V_\cg| \gg |V_\tg|$.
An overview of $\hact$ in Figure~\ref{fig:hierarchical_graph}\textcolor{red}{(c)} displays the multi-level graphs and the hierarchies.

\subsection{HACT Graph Neural Networks (HACT-Net)}
$\hact$-Net processes a multi-scale representation of the tissue. Given $G_\hact$, we learn a graph-level embedding $h_\hact \in \mathbb{R}^{d_\hact}$ that is input to a  classification neural network to predict the classes.
We use the Graph Isomorphism Network ($\gin$)~\cite{xu19}, an instance of message passing neural network~\cite{gilmer17} with a provably strong expressive power to learn fixed-size discriminative graph embeddings.

First, we apply $T_{\cg}$ GIN layers on $G_\cg$ to build contextualised cell-node embeddings. For a node $u$, we iteratively update the node embedding as:

\begin{equation} \label{eq:node_update}
    h_\cg^{(t+1)}(u) = \mathrm{MLP}\Big(h_\cg^{(t)}(u) + \sum_{w \in \mathcal{N}_\cg(u))} h_\cg^{(t)}(w)\Big)
\end{equation}

where, $t=0,\dots,T_{\cg}$, $\mathcal{N}_\cg(u)$ denotes the set of neighborhood cell-nodes of $u$, and MLP is a multi-layer perceptron. At $t=0$, the initial node embedding is, \ie $h_\cg^{(0)}(u) = f_\cg(u)$. 
After $T_\cg$ GIN layers, the node embeddings $\{h^{(T_\cg)}_\cg(u) \; | \; u \in V_\cg\}$ are used as additional tissue-node features, \ie

\begin{align}
    h_\tg^{(0)}(v) = \mathrm{Concat}\Big(f_\tg(v), \sum_{u \in \mathcal{S}(v)} h_\cg^{(T_\cg)}(u)\Big)
\end{align}

where, $\mathcal{S}(v) := \{u \in V_\cg \; | \; S_{\cg \rightarrow \tg}(u, v) = 1 \}$ denotes the set of nodes in $G_\cg$ mapping to a node $v \in V_{\tg}$ in $G_\tg$.
Analogous to Equation~\eqref{eq:node_update}, we apply the second graph neural network based on GIN layers to $G_\tg$ to compute the tissue-node embeddings $\{h^{(t)}_\tg(v) \; | \; v \in V_\tg \}$. At $t=T_\tg$, each tissue-node embeddings encode the cellular and tissue information up to $T_\tg$-hops from $v$. 

% In its local form, the tissue-node update is defined as:
% %
% \begin{align}
%     h_\tg^{(t+1)}(v) = \mathrm{MLP}\Bigg(
%     &\mathrm{Concat}\Big(h_\tg^{(t)}(v), \sum_{w \in \mathcal{S}(v)} h_\cg^{(T_\cg)}(w)\Big) +\\
%     &\sum_{u \in \mathcal{N}_\tg(v)} \mathrm{Concat}\Big(h_\tg^{(t)}(u), \sum_{w \in \mathcal{S}(u))} h_\cg^{(T_\cg)}(w)\Big)
%     \Bigg)
% \end{align}
% %
% where $\mathcal{N}_\tg(v)$ denotes the neighborhood tissue-nodes of $v$ and $\mathcal{S}(v)$ the set of nodes from $\cg$ belonging to the node $v$ of $\tg$. 

Finally, the graph level representation $h_\hact$ is built by concatenating the aggregated node embeddings of $G_\tg$ from all layers~\cite{xu19}, \ie

\begin{equation} \label{eq:concat}
    h_\hact = \mathrm{Concat}\Big(\Big\{\sum_{v \in G_\tg} h_\tg^{(t)}(v) \;\Big|\;t = 0, \dots , T_\tg\Big\} \Big)
\end{equation}

The graph-level representations are then processed by an MLP classifier to predict the cancer subtype.

\section{Experimental Results}
\label{results}
%In this section, we explain the experimental dataset, the implementation of HACT-Net, define the baseline methods for empirical comparison, and present our experimental results.

\subsection{Dataset} 
We introduce a new dataset for BReAst Carcinoma Subtyping (\textbf{BRACS})\footnote{currently pending approval for releasing the dataset to the research community}. 
BRACS consists of 2080 $\roi$s acquired from 106 H\&E stained breast carcinoma whole-slide-images (WSI). The WSIs are scanned with Aperio AT2 scanner at 0.25 $\mu$m/pixel for 40$\times$ resolution.
$\roi$s are selected and annotated as: Normal, Benign (includes Benign and Usual ductal hyperplasia), Atypical (includes Flat epithelial atypia and Atypical ductal hyperplasia), Ductal carcinoma in situ and Invasive, by the consensus of three pathologists using QuPath\cite{bankhead17}. BRACS is more than four times the size of the popular BACH dataset~\cite{aresta19} and consists of challenging typical and atypical hyperplasia subtypes.
Unlike BACH, BRACS exhibits large variability in the $\roi$ dimensions as shown in Table~\ref{tab:stats}. 
The $\roi$s represent a more realistic scenario by including single and multiple glandular regions, and comprising of prominent diagnostic challenges such as stain variance, tissue preparation artifacts and tissue marking artifacts.
Unlike recent graph-based approaches on histopathology data \cite{zhou19, wang19tma, chen19} that conduct data splitting at image level, we perform train, validation and test $\roi$ splits at the WSI-level, such that two images from the same slide does not belong to different splits.
$\roi$s from the same WSI can be morphologically and structurally correlated, even if they are non-overlapping. Thus, image-level splitting leads to over-estimated results on the evaluation set, and networks trained in such manner lack generalizability to unseen data.
We consider four sets of train, validation and test splits, generated at random at the WSI-level, to evaluate our methodology.
% National Cancer Institute - IRCCS-Fondazione Pascale, Italy. 

\begin{table*}
\caption{BReAst Carcinoma Subtyping (BRACS) dataset statistics.}
\label{tab:stats}
\centering
\begin{tabular}{lccccc|c}
\toprule
              & Normal & Benign & Atypical & DCIS & Invasive & Total \\
\midrule
\# RoI   & $305$   & $462$   & $387$    & $503$   & $423$  & $2080$ \\
Avg. \# pixels in a RoI   & $2.1$M  & $5.8$M  & $1.4$M   & $4.4$M  & $9.6$M & $4.9$M \\
Avg. \# nodes in a Cell-graph   & $841$   & $2125$  & $584$   & $1740$   & $4176$ & $1974$ \\
Avg. \# nodes in a Tissue-graph   & $80$    & $222$   & $70$    & $205$    & $487$  & $223$ \\
\end{tabular}

\begin{tabular}{lc|c|c|c|c|c}
\toprule
%\backslashbox{Fold}{Tr,V,Te}
Fold$\sim$Tr/V/Te & Normal & Benign & Atypical & DCIS & Invasive & Total \\
\midrule
Fold 1 & $198/60/47$ & $318/78/66$ & $244/75/68$ & $359/76/68$ & $286/70/67$ & $1405/359/316$ \\
Fold 2 & $202/47/56$ & $304/66/92$ & $245/68/74$ & $355/68/80$ & $283/67/73$ & $1389/316/375$ \\
Fold 3 & $200/56/49$ & $278/92/92$ & $234/74/79$ & $346/80/77$ & $282/73/68$ & $1340/375/365$ \\
Fold 4 & $196/49/60$ & $292/92/78$ & $233/79/75$ & $350/77/76$ & $285/68/70$ & $1356/365/359$ \\
\bottomrule
\end{tabular}
\end{table*}

\begin{figure*}
\centering
% normal 
\begin{subfigure}[t]{.25\textwidth}
\centering
\includegraphics[width=\linewidth]{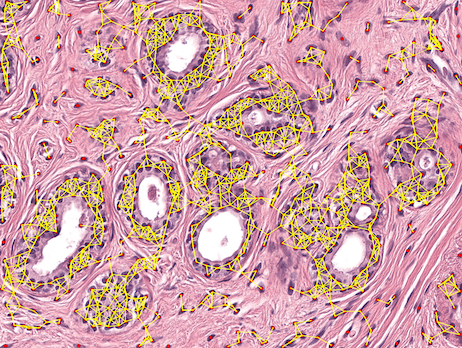}
\captionsetup{width=\textwidth}
\caption{}
\end{subfigure}%
\begin{subfigure}[t]{.25\textwidth} 
\centering
\includegraphics[width=\linewidth]{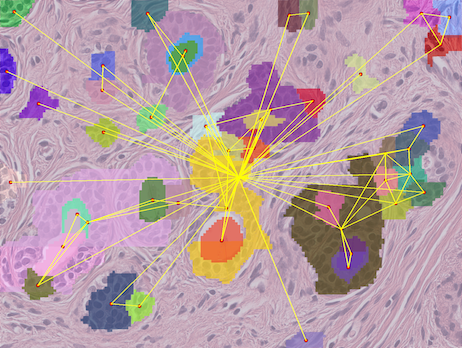}
\captionsetup{width=\textwidth}
\caption{}
\end{subfigure}%

% Benign
\begin{subfigure}[t]{.25\textwidth}
\centering
\includegraphics[width=\linewidth]{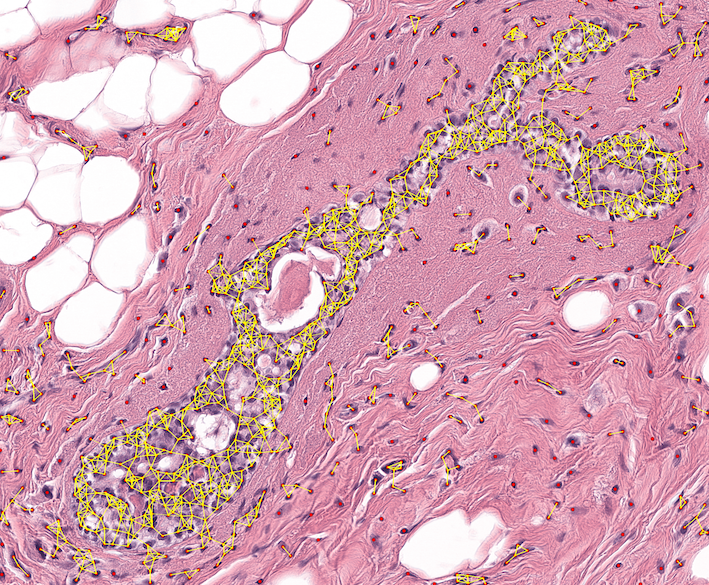}
\captionsetup{width=\textwidth}
\caption{}
\end{subfigure}%
\begin{subfigure}[t]{.25\textwidth} 
\centering
\includegraphics[width=\linewidth]{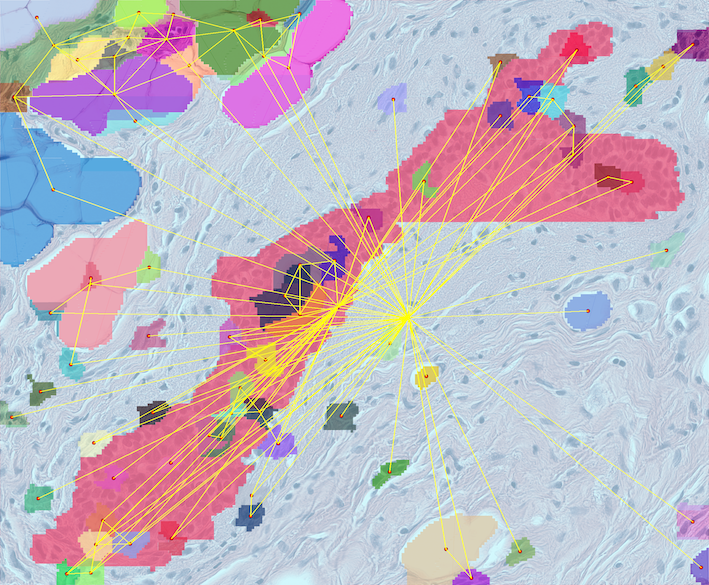}
\captionsetup{width=\textwidth}
\caption{}
\end{subfigure}%

%\end{figure}
%\begin{figure}[ht]\ContinuedFloat

\centering
% Atypical
\begin{subfigure}[t]{.25\textwidth} 
\centering
\includegraphics[width=\linewidth]{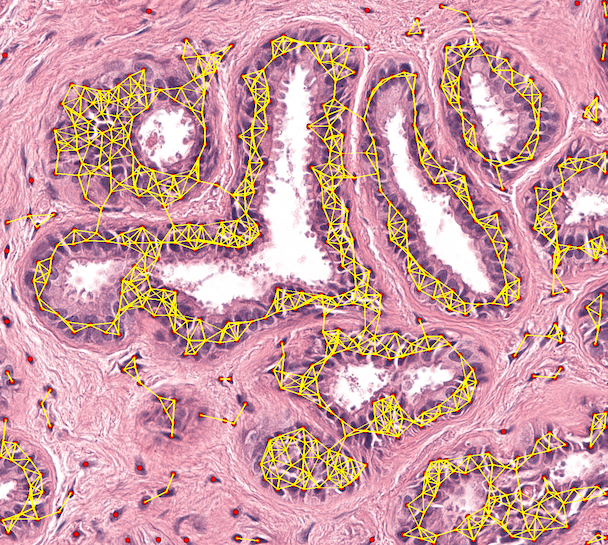}
\captionsetup{width=\textwidth}
\caption{}
\end{subfigure}%
\begin{subfigure}[t]{.25\textwidth} 
\centering
\includegraphics[width=\linewidth]{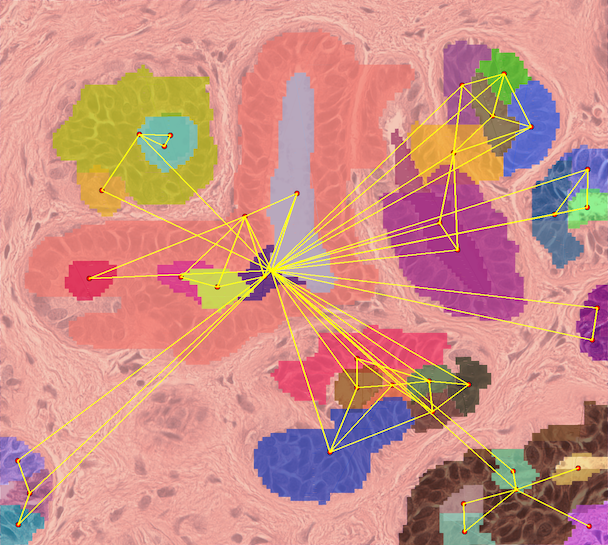}
\captionsetup{width=\textwidth}
\caption{}
\end{subfigure}%

% DCIS
\begin{subfigure}[t]{.25\textwidth} 
\centering
\includegraphics[width=\linewidth]{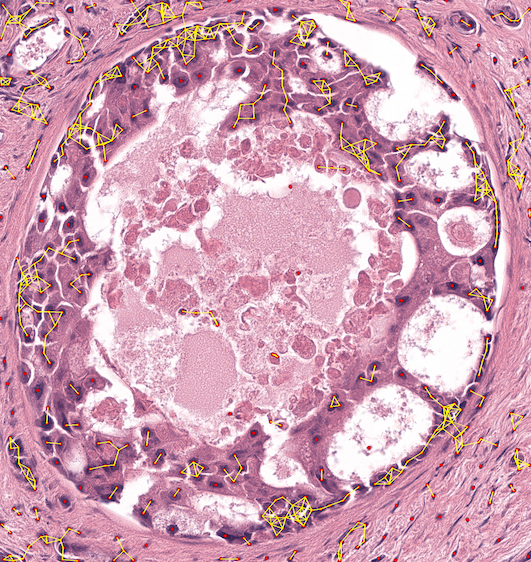}
\captionsetup{width=\textwidth}
\caption{}
\end{subfigure}
\begin{subfigure}[t]{.25\textwidth} 
\centering
\includegraphics[width=\linewidth]{figures/1238_dcis_15_tissue_graph.png}
\captionsetup{width=\textwidth}
\caption{}
\end{subfigure}

% Invasive 
\begin{subfigure}[t]{.30\textwidth} 
\centering
\includegraphics[width=\linewidth]{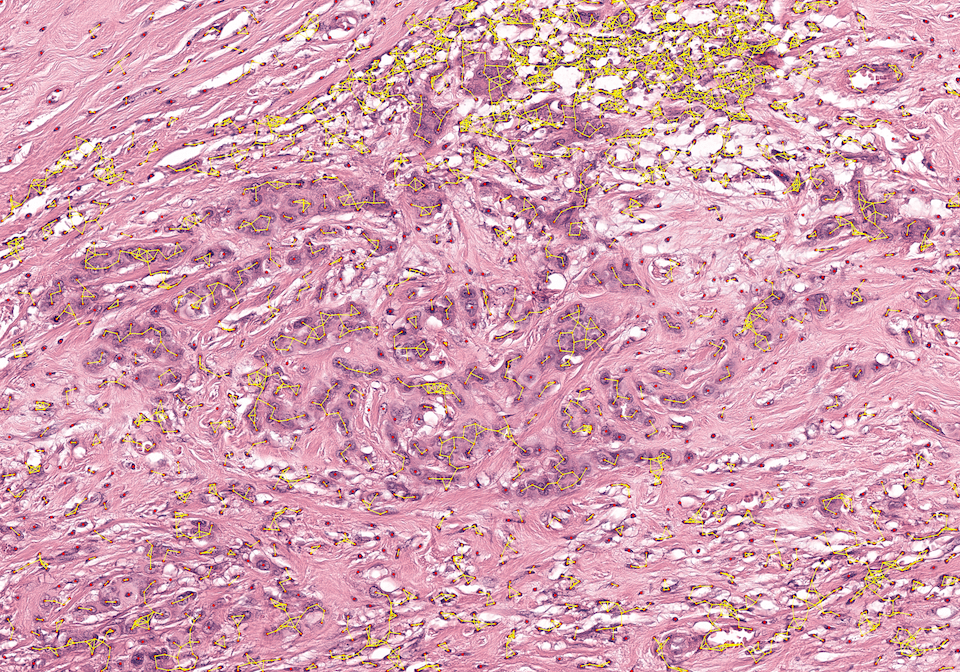}
\captionsetup{width=\textwidth}
\caption{}
\end{subfigure}
\begin{subfigure}[t]{.30\textwidth} 
\centering
\includegraphics[width=\linewidth]{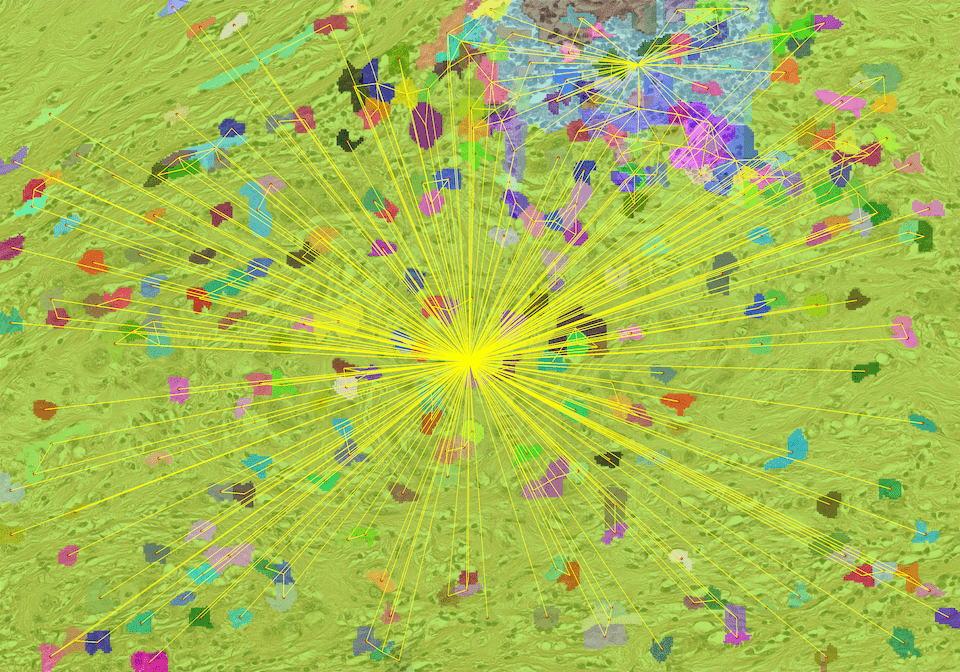}
\captionsetup{width=\textwidth}
\caption{}
\end{subfigure}

\caption{Cell-graph (left) and tissue-graph (right) examples for four cancer subtypes. (a-b) Normal, (c-d) Benign, (e-f) Atypical, (g-h) DCIS, and (i-j) Invasive. Large central nodes in the tissue-graphs depict the centroids of the surrounding stroma tissues.}

\label{fig:examples}
\end{figure*}

\begin{table*}
\caption{Weighted F1-scores across four test folds. Mean and standard deviation of fold-wise and class-wise weighted F1-scores. Results expressed in $\%$.}
\label{tab:results}
\centering
% \setlength{\belowcaptionskip}{-5pt}
% \setlength{\tabcolsep}{0.5em}
% \renewcommand{\arraystretch}{1.2}
% \scriptsize 
\begin{tabular}{lccccc|ccccc}
  \toprule
    Model/Fold\# & 1 & 2 & 3 & 4 & $\mu\pm\sigma$ & Normal & Benign & Atypical & DCIS & Invasive \\
    \midrule
    $\cnn~(10\times)$~\cite{sirinukunwattana18,mercan19} & $49.85$ & $46.86$ & $51.19$ & $54.04$ & $50.49$ & $47.50$ & $46.00$ & $39.25$ & $51.25$ & $69.75$\\
                             & & & & & $\pm2.58$ & $\pm2.50$ & $\pm7.45$ & $\pm3.63$ & $\pm2.05$ & $\pm4.21$  \\[0.1cm]
                             
    $\cnn~(20\times)$~\cite{sirinukunwattana18,mercan19} & $52.49$ & $51.88$ & $44.38$ & $56.37$ & $51.28$ & $52.25$ & $47.25$ & $44.50$ & $48.25$ & $62.25$\\
                             & & & & & $\pm4.34$ & $\pm1.64$ & $\pm6.80$ & $\pm4.56$ & $\pm3.56$ & $\pm4.44$  \\[0.1cm]
                             
    $\cnn~(40\times)$~\cite{sirinukunwattana18,mercan19} & $40.64$ & $47.30$ & $38.08$ & $48.95$ & $43.74$ & $46.00$ & $35.50$ & $46.75$ & $38.00$ & $56.00$\\
                             & & & & & $\pm4.51$ & $\pm7.71$ & $\pm8.96$ & $\pm5.02$ & $\pm4.30$ & $\pm7.12$  \\[0.1cm]
                             
    Multi-scale $\cnn$ & $56.17$ & $54.41$ & $53.94$ & $55.66$ & $55.04$ & $57.25$ & $51.75$ & $42.25$ & $54.50$ & $72.25$\\
   ($10\times$+$20\times$)~\cite{sirinukunwattana18,mercan19} & & & & & $\pm0.90$ & $\pm3.90$ & $\pm8.78$ & $\pm8.73$ & $\pm2.06$ & $\pm1.92$  \\[0.1cm]
   
    Multi-scale $\cnn$ & $58.80$ & $54.64$ & $55.53$ & $53.90$ & $55.72$ & $55.75$ & $52.25$ & $46.75$ & $50.75$ & $71.75$\\
   ($10\times$+$20\times$+$40\times$)~\cite{sirinukunwattana18,mercan19} & & & & & $\pm1.87$ & $\pm1.78$ & $\pm6.38$ & $\pm2.28$ & $\pm2.38$ & $\pm3.34$  \\[0.1cm]

    %$\cnn~(20\times)$~\cite{add-ref}       & $59.12$       & $58.38$       & $55.33$       & $58.26$       & $57.77$       & $62.00$       & $52.00$       & $46.50$       & $59.75$       & $70.00$\\
    %                        &               &               &               &               & $\pm1.45$    & $\pm3.67$      & $\pm4.30$     & $\pm5.22$     & $\pm4.92$     & $\pm3.94$  \\
    
    %$\cnn~(40\times)$~\cite{add-ref}       & $55.21$       & $53.38$       & $47.97$       & $57.20$       & $53.44$       & $64.75$      & $50.75$      & $39.00$      & $50.25$      & $67.25$ \\
    %&               &               &               &               & $\pm 3.44$    & $\pm 8.17$          & $\pm 4.71$          & $\pm 4.30$          & $\pm 3.90$          & $\pm 4.49$  \\
    
    $\cgc$~\cite{zhou19}       & $51.54$       & $58.97$       & $56.70$       & $50.44$       & $54.41$       & $53.00$      & $52.25$      & $42.00$      & $57.00$      & $68.25$ \\
    &               &               &               &               & $\pm 3.53$    & $\pm 2.55$          & $\pm 4.96$          & $\pm 8.15$          & $\pm 5.52$          & $\pm 2.58$  \\
    \midrule
    
    $\tg$-$\gnn$        & $54.47$       & $55.13$       & $67.84$       & $49.85$       & $56.82$       & $56.78$       & $54.76$       & $48.52$       & $56.53$       & $69.52$ \\
    &               &               &               &               & $\pm 6.67$    & $\pm 1.89$          & $\pm 6.62$          & $\pm 8.76$          & $\pm 12.78$          & $\pm 11.00$  \\[0.1cm]
    
    $\cg$-$\gnn$        & $61.35$       & $53.81$       & $62.00$       & $55.38$       & $58.13$       & $62.66$       & $\mathbf{64.57}$      & $36.18$       & $59.98$       & $68.12$ \\
    &               &               &               &               & $\pm 3.59$    & $\pm 5.32$          & $\pm 9.05$          & $\pm 6.85$          & $\pm 1.43$          & $\pm 2.52$  \\[0.1cm]
    
    $\concat$-$\gnn$        & $54.66$       & $54.49$       & $64.59$       & $\mathbf{63.95}$      & $59.42$       & $57.00$       & $60.31$       & $49.62$     & $60.65$       & $68.94$ \\
    &               &               &               &               & $\pm 4.85$    & $\pm 4.06$          & $\pm 8.36$          & $\pm 4.71$          & $\pm 4.94$          & $\pm 12.47$  \\[0.1cm]
    
    $\hact$-Net     & $\mathbf{62.17}$      & $\mathbf{59.06}$      & $\mathbf{69.41}$      & $60.92$       & $\mathbf{62.89}$       & $\mathbf{65.15}$      & $58.40$        &  $\mathbf{55.45}$     & $\mathbf{63.15}$     & $\mathbf{73.78}$ \\
    &               &               &               &               & $\pm 3.92$    & $\pm 3.64$          & $\pm 10.59$          & $\pm 5.19$          & $\pm 4.08$          & $\pm 7.35$  \\
    \bottomrule
\end{tabular}
\end{table*}

\subsection{Implementation}
All our experiments are conducted using PyTorch~\cite{paszke19} and the DGL library~\cite{wang19}. We benchmark our proposed method, $\hact$-Net, against several $\gnn$- and $\cnn$-based approaches. 
We compare $\hact$-Net with standalone $\cg$-$\gnn$ and $\tg$-$\gnn$ to assess the impact of multi-level information processing.
We compare $\hact$-Net with Concat-$\gnn$ that concatenates the $\cg$ and $\tg$ graph embeddings, \ie $h_\concat = \concat(h_\cg, h_\tg)$, to evaluate the benefit of hierarchical-graph learning. Note that Concat-$\gnn$ is analogous to the recently proposed Pathomic Fusion by~\cite{chen19}.
For the $\cnn$ approaches, we implement single scale $\cnn$s \cite{sirinukunwattana18} at three magnifications. Further, we compare with two multi-scale $\cnn$s utilizing late fusion with single stream + LSTM architecture \cite{sirinukunwattana18}. The multi-scale $\cnn$s use multi-scale patch information from (10$\times$ + 20$\times$) and (10$\times$ + 20$\times$ + 40$\times$). 
Considering tumor heterogeneity, $\cnn$ approaches are limited to 10$\times$ magnification so that only one cancer type is included in an $\roi$. 

% Network architectures
The $\cg$-$\gnn$ and $\tg$-$\gnn$ have four $\gin$ layers with a hidden dimension of $32$ in standalone, Concat-$\gnn$ and $\hact$-Net. 
Each GIN layer uses a 2-layer MLP with $\mathrm{ReLU}$ activation.
% In the GIN layers, we replace the sum aggregator with the mean aggregator.
The classifier is composed of a 2-layer MLP with $64$ hidden neurons and five output neurons, \ie the number of classes. The model is trained to minimize the cross-entropy loss between the output logits and the ground truth labels. 
We set the batch size to $16$, the initial learning rate to $10^{-3}$ and use the Adam~\cite{kingma15} optimizer with a weight decay of $5.10^{-4}$.
For the single-scale and multi-scale $\cnn$s, we extract patches of size 128$\times$128 at $10\times$, $20\times$ and $40\times$. Pre-trained ResNet-50 on ImageNet is finetuned to obtain patch-level feature representations after experimenting with different ResNet, VGG-Net and DenseNet architectures. 
All the $\cnn$s use \cite{mercan19} to derive $\roi$-level feature representation via  aggregate-penultimate technique, and employ a 2-layer MLP with $64$ hidden neurons and five output neurons for $\roi$ classification.
Considering the per-class data imbalance, weighted F1-score is used to quantify the classification performance.
Model with the best weighted F1-score on the validation set is selected as the final model in each approach.

\subsection{Discussion}
Figure \ref{fig:examples} demonstrates $\cg$ and $\tg$ representation of sample $roi$s from BRACS dataset. Visual inspection signifies that the constructed $\cg$s aptly encompass the cellular distribution and cellular interactions. Similarly, the $\tg$s aptly encode the tissue microenvironment by including the topological distribution of the tissue components. The $\tg$s include lumen in Benign, apical snouts in Atypical, necrosis in DCIS and tumor-associated stroma in DCIS and Invasive that are not accessible to the $\cg$s.

Table~\ref{tab:results} presents the weighted F1-score on four test folds and their aggregate statistics for the networks.
The standalone $\cnn$s perform better while operating at lower magnification as they capture larger context. The multi-scale $\cnn$s perform better by including local and global context information from multiple magnifications.
The $\cg$-$\gnn$ and $\tg$-$\gnn$ results signify that topological entity-based paradigm is superior to pixel-based $\cnn$s. Further, they indicate that tissue distribution information is inferior to nuclei distribution information for breast cancer subtyping. 
Our $\cg$-$\gnn$ baseline outperforms $\cgc$~\cite{zhou19} justifying the use of expressive backbone GNNs like $\gin$~\cite{xu19}. We also hypothesize that simply concatenating the updated node representation at \emph{each} layer as shown in Equation~\ref{eq:concat} brings a performance boost without additional parameters. 
$\concat$-$\gnn$ outperforms $\tg$-$\gnn$ and $\cg$-$\gnn$ indicating that $\cg$ and $\tg$ provide valuable complementary information. 
Further, $\hact$-Net outperforms $\concat$-$\gnn$ confirming that the \emph{relationship} between the low and high-level information must be modeled at the local node-level rather than at the graph-level for better structure-function mapping.

The class-wise performance analysis in Table~\ref{tab:results} shows that invasive category is the best detected. It translates to the topologically recognizable patterns with scattered nodes and edges in $\cg$ and $\tg$. 
Atypical cases are the hardest to model, partially as they have a high intra-class variability and high inter-class ambiguity with benign and DCIS. Large drops in performance in the CGCNet and $\cg$-$\gnn$ for the atypical category convey that the standalone cell information is not discriminative enough to identify these patterns. Tissue information such as apical snouts in FEA, necrosis in DCIS, stroma microenvironment in Benign etc. bolster the discriminability of atypical $\roi$s. Thus, all the networks including $\tg$ perform better than $\cg$-$\gnn$ for the atypical category.
The $\cg$-$\gnn$ and $\tg$-$\gnn$ performances for the Normal, Benign and DCIS indicate that nuclei information is more informative to identify these categories. 
$\hact$-Net utilizes both nuclei and tissue distribution properties, thus performing superior to $\cg$-$\gnn$ and $\tg$-$\gnn$ for almost all subtypes. Unlike $\cg$-$\gnn$, $\hact$-Net utilizes stromal microenvironment around the tumor regions which is a pivotal factor in breast cancer development \cite{bejnordi18}.
The class-wise comparison between $\hact$-Net and $\concat$-$\gnn$ establish the positive impact of hierarchical learning.
The gain in class-wise performances of $\hact$-Net substantiates that the network does not get biased towards one particular class.

Moreover, the paradigm shift from pixel-based analysis to entity-based analysis can potentially yield interpretability of the deep learning techniques in digital pathology. For instance, \cite{zhou19} analyzes the cluster assignment of each node in $\cg$ representation to conclude that the clustering operation groups cells according to their appearance and tissue belongingness. \cite{jaume20} introduced a novel post-hoc interpretability module on top of the learned $\cg$-$\gnn$ to identify decisive sets of cells and cellular interactions. However, both approaches are limited to $\cg$ analysis.
Since $\hact$-representation captures entity-based multi-level hierarchical tissue attributes similar to pathological diagnostic procedure, the interpretability of $\hact$-representation can  identify crucial entities, such as nuclei, tissue parts and cell-to-tissue interactions, to imitate the pathologist's assessment.

\section{Conclusion}
\label{conclusion}
In this work, we have proposed a novel hierarchical tissue representation in combination with a hierarchical $\gnn$ to map the histopathological structure to function relationship.
We have extensively evaluated the proposed methodology and compared with the state-of-the-art $\cnn$s and $\gnn$s for breast cancer subtyping.
The enriched multi-level topological representation and hierarchical learning scheme strengthens the proposed methodology to result in superior classification performance.
The $\hact$-representation can seamlessly scale to any sized $\roi$ to incorporate local and global context for improved stratification.
The entity-based graphical representation yields better control for tissue encoding, and favors the inclusion of pathological context into the modeling. The success of our methodology inspires to explore approaches to further include pathological priors.
Further, the hierarchical modeling paves way for recent interpretability techniques in digital pathology to go beyond cell-graphs to interpret the hierarchical nature of the tissue. 

\bibliography{main}
\bibliographystyle{IEEEtran}

\end{document}